\documentclass{article}

\usepackage{PRIMEarxiv}

\usepackage[utf8]{inputenc} 
\usepackage[T1]{fontenc}    
\usepackage{hyperref}       
\usepackage{url}            
\usepackage{booktabs}       
\usepackage{amsfonts}       
\usepackage{nicefrac}       
\usepackage{microtype}      
\usepackage{fancyhdr}       
\usepackage{graphicx}       
\usepackage{xcolor}
\usepackage{mdframed}
\usepackage{amssymb} 
\usepackage{pifont} 
\newcommand{\cmark}{\ding{51}}%
\newcommand{\xmark}{\ding{55}}%
\usepackage{enumitem}

\graphicspath{{media/}}     
\newmdenv[
    backgroundcolor=gray!10,    
    linecolor=blue!75!black,    
    innerleftmargin=10pt,       
    innerrightmargin=10pt,      
    innertopmargin=10pt,        
    innerbottommargin=10pt,     
    roundcorner=5pt,            
]{custommdframed}

\definecolor{softblue}{RGB}{183, 207, 237} 
\definecolor{lightblue}{RGB}{232, 236, 248} 

\newmdenv[
    backgroundcolor=gray!10,    
    linecolor=blue!75!black,    
    innerleftmargin=10pt,
    innerrightmargin=10pt,
    innertopmargin=10pt,
    innerbottommargin=10pt,
    roundcorner=5pt,
]{questionmdframed}
\title{A template for Arxiv Style
\thanks{\textit{\underline{Citation}}: 
\textbf{Authors. Title. Pages.... DOI:000000/11111.}} 
}

\usepackage[utf8]{inputenc}
\usepackage{hyperref}
\usepackage{graphicx}
\usepackage{amsmath}
\usepackage{amsfonts}
\usepackage[numbers]{natbib}
\title{LLM Questionnaire Completion for Automatic Psychiatric Assessment}
\author{
  Gony Rosenman \\
  Sagol School of Neuroscience \\
  Tel Aviv University \\
  \And
  Lior Wolf \\
  School of Computer Science \\
  Tel Aviv University \\
  \And
  Talma Hendler \\
  Sagol School of Neuroscience \\
  Tel Aviv University \\
}

\date{\today}

\begin{document}
\maketitle

\begin{abstract} 
We employ a Large Language Model (LLM) to convert unstructured psychological interviews into structured questionnaires spanning various psychiatric and personality domains. The LLM is prompted to answer these questionnaires by impersonating the interviewee. The obtained answers are coded as features, which are used to predict standardized psychiatric measures of depression (PHQ-8) and PTSD (PCL-C), using a Random Forest regressor. Our approach is shown to enhance diagnostic accuracy compared to multiple baselines. It thus establishes a novel framework for interpreting unstructured psychological interviews, bridging the gap between narrative-driven and data-driven approaches for mental health assessment. 
\end{abstract}

\section{Introduction}
Psychiatric evaluation nowadays is heavily dependent on the patient's verbal report about disturbed feelings, thoughts, behaviors and their changes over time. Accordingly, evaluation hinges on two main components: unstructured interviews, which allow patients to express themselves freely under the guidance of open questions, and structured questionnaires, aimed at standardizing the assessment. These methods are outlined in the Diagnostic and Statistical Manual of Mental Disorders (DSM) series, which attempts to assign universal scores to individual experiences of mental disorders \cite{APA2013DSM5}. However, the inherent complexity of mental health conditions, characterized by a known positive manifold of symptoms and compounded by the subjective nature and potential unreliability of self-reported data (especially from one session to another), along with interviewer biases, make accurate diagnosis challenging. The overlapping symptoms and the instability of mental state, especially in pathological conditions, further complicate the need for precision, precluding an objective and quantitative account of a critical element in the psychiatric evaluation process; the subjective self-experience \cite{Vanes2021,Dwyer2018,Bzdok2018}.

The evolution of psychiatric practice is increasingly shaped by the integration of Natural Language Processing (NLP) and machine learning within traditional diagnostic approaches. This shift signifies a move from a predominantly theory-driven to a more data-driven paradigm, addressing limitations in conventional psychiatric assessments. NLP's role is pivotal in transcending the subjective interpretation of unstructured interviews and the rigidity of standardized questionnaires. Leveraging large-scale textual data and advanced language models, NLP enriches psychiatric evaluations with nuanced, patient-specific insights. This data-driven approach, complementing established psychiatric theories, fosters a more holistic and precise understanding of mental health conditions, aligning with the trend towards precision medicine in mental healthcare.

We introduce a two-step method for psychiatric evaluation based on unstructured interview text. First, we ask the LLM to fill in multiple questionnaires while impersonating the interviewee. These questionnaires are of two types: (i) established psychiatric questionnaires (PHQ-8, PCL-C), and (ii) questionnaires we developed using GPT-4, which cover mental health issues, personality traits, and therapeutic dimensions. In the second step, the response of the LLM is used directly, as features for a Random Forest Regressor~\cite{Breiman2001} that is used to predict the interviewee scores on the two clinical questionnaires. 

\section{Related Work}
Earlier diagnostic efforts in mental health\cite{Asgari2014,alhanai18_interspeech,Niu2021,Dai2021,Lu2022}, utilize both the textual information and the speech intonation features, acknowledging the importance of tonal elements in depression assessment. \citet{Sun2017} manually extract six psychological dimensions (sleep quality, existing PTSD/Depression diagnostic, treatment history, introversion, personal preference, and feeling) from interview text and then employ a Random Forest regressor.

More recently, the adoption of transformer architectures has brought significant advancements for depression diagnosis. \citet{Milintsevich2023} utilize the RoBERTa transformer~\cite{liu2019roberta} for encoding interview segments, focusing on predicting individual depression symptoms. \citet{zhang2024multilevel} introduces prompt learning and an attention mechanism to improve symptom prediction. These contributions rely on supervised learning for predicting questionnaires, while our work utilizes zero-shot impersonation. 

Recently, \citet{GalatzerLevy2023TheCO} explored the use of a medical knowledge-enhanced LLM Med-PaLM 2~\cite{singhal2023towards}, for psychiatric prediction. This effort is aimed not at answering the questionnaire, but at directly predicting the outcome using the prompt ``Based on the following clinical interview, what do you estimate the Participant's 
[PHQ-8/PCL-C] score is?''. Concurrently, \cite{Mentallama} runs prompt-based few-shot learning~\cite{brown2020language} to analyze social media content and directly answers questions such as ``What mental disorder
symptoms does this post show?'', focusing on the quality of the textual explanation within the LLM's output. Similarly, \citet{Sadeghi2023} utilized a combination of LLM summarization prompts and a fine-tuned transformer model for depression detection, highlighting the evolving utility of language models in clinical diagnostics.

\begin{figure*}[t]
    \centering
    \includegraphics[width=\textwidth]{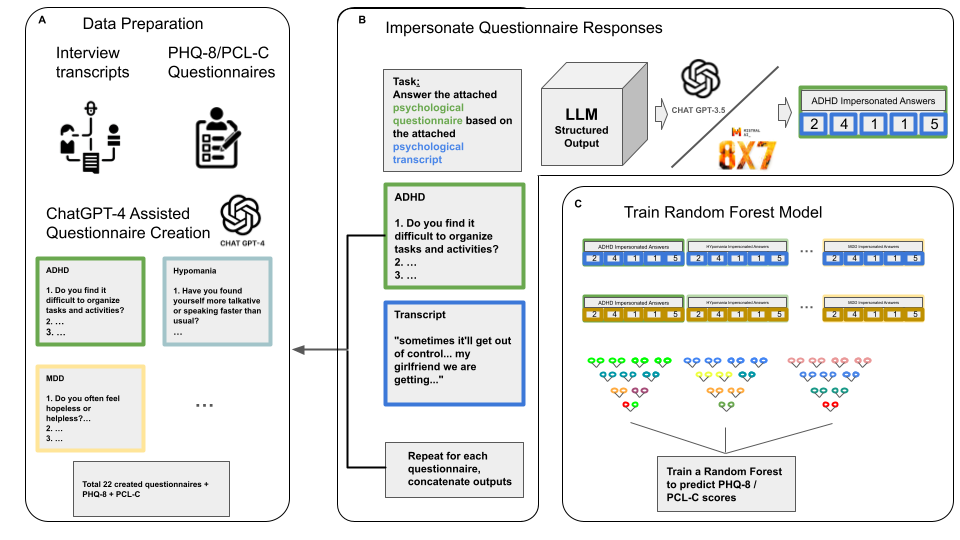}
    \caption{LMIQ Pipeline Overview. A. Top. Preparation of the Extended-Daic dataset including 275 subjects with psychological interview transcripts, along with PTSD and Depression scores derived from PCL-C and PHQ-8 assessments. Bottom. Instruction for Chat-GPT 4 to develop five-item questionnaires spanning multiple mental health and personality domains. B. Main Prompt Logic. Merging of a task description with a psychological interview transcript and a questionnaire to generate five impersonated responses. Aggregate across all subjects and questionnaires. C. Training of a Random Forest model using the questionnaire responses ($d=135$) to accurately predict the original assessment scores.}
    \label{fig:overview}
\end{figure*}

\section{Methodology}
This work leverages the Extended Distress Analysis Interview Corpus (E-DAIC) \cite{edaic}, a dataset of semi-clinical interviews designed to assess psychological distress.
To build the corpus, an autonomous AI interviewer was employed in order to minimize human bias, and two clinical questionnaires-based scales, PHQ-8 and PCL-C, were used to measure both depression and Post Traumatic Stress Disorder (PTSD). 
Our analyses adhere to the predefined training, development, and test splits specified by the dataset's documentation. 
The E-DAIC dataset comprises $n=275$ psychological transcripts, each associated with two psychiatric scores, which quantify the severity of depression and PTSD, measured using the PHQ-8 and the PCL-C standardized scales, respectively.

Given a text in the domain of psychological transcripts $t_i\in \text{PT}$, where $i \in \{1, 2, \ldots, 275\}$, our objective is to design a feature extractor $F: \text{PT} \to \mathbb{R}^d$. This extractor converts a given transcript $t_i$ into a $d$-dimensional feature vector $\mathbf{v}_i$ such that the two ground truth scores $s_i = [s^\text{DEPRESSION}_i,s^\text{PTSD}_i]$ can be inferred accurately. Upon extracting these features, our next step involves employing a Random Forest Regressor, $R: \mathbb{R}^d \to \mathbb{R}^2$, which takes as input the feature vector $\mathbf{v}_i$ and predicts the pair of psychiatric scores $s_i$.

Our method, Language Model for Impersonation-based Questionnaire Completion (LMIQ), defines a feature extractor \( F \) as the transformation of unstructured interview transcripts into a structured array of impersonated responses to a list of questions. For each subject and question, the method: (1) generates a query by combining the transcript with a psychological questionnaire, (2) processes the query through an LLM to obtain responses, and (3) stores the responses in an array, which represents the feature vector for that subject.

The system prompt directed the LLM to interpret transcripts and simulate responses as the interview subject. The exact prompt used in our research is detailed in Appendix~\ref{app:prompt}, and a shorter reference prompt is as follows.
\smallskip
\begin{custommdframed}
Attached are a psychological interview transcript and psychological questionnaire. \\
Analyze the conversation transcript and capture psychological insights about the speaker. \\
Deduct answers to the accompanied questions as if you were the speaker. \\
The answer should be a numerical value ranging from 1-5, with 1 depicting ``don't agree at all'' and 5 depicting ``very much agree.''\\
Conversation Transcript: \{transcript\} \\
Psychological Questionnaire: \{question\}
\end{custommdframed}

In our implementation, LMIQ utilizes GPT-3.5 Turbo Instruct, via the OpenAI API (\href{https://platform.openai.com/docs/models/gpt-3-5}{https://platform.openai.com/docs/models/gpt-3-5}) for transcript analysis and predicting the responses to psychological questionnaires. Additionally, we evaluated LMIQ that utilizes Mixtral 7Bx8 \cite{Mixtral2024} as its LLM backbone. LLMs were used with their default configuration, including temperature settings, to maintain consistent and unbiased processing. The individual questions in the questionnaires are submitted to the LLM one by one in separate sessions, to avoid biases that could be caused by their order.

The list of questions employed by LMIQ includes the original eight-item PHQ-8 and seventeen-item PCL-C questionnaires, as well as multiple questionnaires we developed using OpenAI's ChatGPT-4. We prompted it to create multiple five-item sets across three domains: (i) clinical mental health (DSM-5 Guided), (ii) the five-factor model of personality, and (iii) therapeutic relevant domains. The inclusion of a wider array of questionnaires beyond the PHQ-8 and PCL-C is motivated by the concept of psychiatric comorbidity and its symptom overlap.

Within each domain, GPT-4 was prompted to define multiple topics and within each topic to generate exactly five questions, each aiming to quantify symptoms on a scale between 1 and 5.  These initial outputs were manually refined to ensure higher relevance and precision. The obtained domains are listed in Tab.~\ref{table:domains}, the full list of questions is provided in Appendix~\ref{app:questions}; as a sample, the generalized anxiety disorder questions include: (1) Do you find it hard to control your worrying? and (2) Does your anxiety interfere with your work, school, or family responsibilities?

\begin{table}[t]
\centering
\begin{tabular}{@{}lp{4cm}p{1cm}@{}}
\toprule
\textbf{Domain} & \textbf{Questionnaires} & \textbf{\#Ques-tions} \\ \midrule
Mental health & Agoraphobia, ADHD, Body Image, Borderline Personality Disorder, etc.
& 70 \\ 
Personality & Agreeableness, Conscientiousness, Extraversion, Neuroticism, Openness & 25 \\ 
Therapeutic & Family History, Trauma History, Resilience & 15 \\ 
Direct & PHQ-8, PCL-C & 25\\ \bottomrule
\end{tabular}
\caption{Questionnaire Domains and Constituents}
\label{table:domains}
\end{table}
\begin{table*}[t]
\centering
\begin{tabular}{lcccc}
\toprule
& \multicolumn{2}{c}{PHQ depression score} & \multicolumn{2}{c}{PTSD severity scale} \\
\cmidrule(lr){2-3}
\cmidrule(lr){4-5}
{Model} & {Dev MSE} & {Test MSE} & {Dev MSE} & {Test MSE} \\
\midrule
LMIQ & 23.87 & 20.42 & 144.17 & 192.93 \\
LMIQ (Mixtral) &  18.50 &  18.05 &  90.11 &  163.75 \\
GPT-3.5 Impersonate Zero-Shot & 27.62 & 24.16 & 154.66 & 195.55 \\
GPT-3.5 Zero-Shot & 35.62 & 33.42 & 251.70 & 336.49 \\
GPT-3.5 Analyze \& Embed & 29.81 & 35.80 & 216.13 & 348.02 \\
Direct Embedding with Ada-002 & 31.83 & 38.23 & 226.18 & 389.65 \\
MentaLLama 7B Analyze \& Embed & 29.61 & 39.40 & 216.30 & 367.09 \\
Guessing the mean value & 32.06 & 42.80 & 234.52 & 407.2 \\
TF-IDF Vectorization & 31.47 & 43.36 & 227.61 & 392.63 \\
\bottomrule
\end{tabular}
\caption{Summary of Model Performances in PHQ-Score and PTSD Severity Prediction Tasks}
\label{tab:performace}
\vspace{-.3cm}
\end{table*}

\begin{table}
\centering
\begin{tabular}{@{}l@{~~}c@{~~}c@{}}
\toprule
& {MAE} & {RMSE} \\ 
{Method} & Dev/Test & Dev/Test  \\ \midrule
LMIQ (Ours) & 3.87/3.46 & 4.78/4.30 \\ 
\citet{zhang2024multilevel} & -/ 5.28 & - / - \\ 
\citet{Sadeghi2023} & 3.65/4.26 & 5.27/5.36 \\
\citet{Niu2021} & 3.73 / - & 4.80 / - \\ 
\citet{Dai2021} & 3.22/3.98 & 4.43/5.11 \\
\citet{Lu2022} & 4.48/- & 5.37/- \\
\citet{Milintsevich2023} & 5.51/ 5.03 & -/ - \\ \bottomrule
\end{tabular}
\caption{Performance on the DAIC-WoZ Dataset}
\label{table:daic_woz_mae_rmse_comparison}
\vspace{-.5cm}
\end{table}

The feature vector was formed by concatenating responses (each an integer between one and five) to five questions across twenty-four topics, comprising a vector with $d=135$ dimensions. This vector was then used as input for a Random Forest Regressor implemented using Scikit-Learn \cite{scikitlearn}. Model optimization was conducted through a hyper-parameter search, specifically examining 'n-estimators': [100, 200, 300] and 'max-depth': [10, 20, 30]. The configuration that yielded the best performance on the development set of E-DAIC was subsequently applied to the test set for evaluation. A comprehensive overview of our methodology, from data collection to prediction, is detailed in Figure~\ref{fig:overview}.

\section{Experiments}
We employ multiple appropriate baselines and various ablated versions of our method to demonstrate the advantage of our approach in the context of predicting psychiatric scores from psychological transcripts. Unless specified otherwise, all feature extraction methods are coupled with the same Random Forest Regression that is used for LMIQ.

{\bf Direct Embedding with Ada-002}
This baseline method utilizes text-embedding-ada-002 \cite{ada002}, a model from the OpenAI API, to transform psychological transcripts into vector embeddings.

{\bf GPT-3.5 Analyze \& Embed}
This baseline combines GPT-3.5's analysis of psychological transcripts, with a focus on conditions such as depression and PTSD, and a subsequent text-embedding-ada-002 of the response. 
The prompt used and a sample response are provided in Appendix~\ref{app:prompta}.

{\bf MentalLLAMA 7B  Analyze \& Embed}
Similar to the GPT-3.5  Analyze \& Embed, but utilizing a MentaLLama 7B model \cite{Mentallama}, a LLAMA \cite{llama} model that was fine-tuned for depression detection on a large dataset of Reddit posts in a mental health sub-Reddit. See Appendix~\ref{app:promptml} for the sample output. 

{\bf TF-IDF Vectorization}
This approach represents a more traditional text processing technique. Each transcript is vectorized using the Term Frequency-Inverse Document Frequency (TF-IDF) method. 

{\bf GPT-3.5 Analyze \& Predict}
This baseline directs GPT-3.5 to perform a zero-shot prediction of psychiatric scores PHQ-8 and PCL-C, similarly to~\cite{GalatzerLevy2023TheCO}. 

{\bf GPT-3.5 Impersonate Zero-Shot}
This baseline instructs GPT-3.5 to answer the PHQ-8 and PCL-C questionnaires as if it were the subject, based on the psychological transcript. This method directly calculates scores as the sum of questionnaire responses, mirroring clinical psychiatric evaluation. Differing from the main method, it bypasses the Random Forest model and operates as a zero-shot approach.

{\bf Naive Average Guess}
To provide a scale regarding the variability of the scores, the mean value of the training set is used as the prediction for all subjects regardless of the psychological transcript.

\section{Results and Discussion}
The results in Table~\ref{tab:performace} indicate that the LMIQ model exhibits lower Mean Squared Errors (MSEs) than all baselines for both depression and PTSD severity, indicating superior accuracy. 
Additionally, LMIQ, utilizing the Mixtral 7Bx8 model—which is believed to surpass GPT-3.5 in performance—indeed demonstrates improved outcomes. This underscores the effectiveness of our approach and suggests that its potential is amplified with advancements in LLMs.
The next best method is the one that predicts the clinical questionnaires (but not other questionnaires) and sums the results instead of using a regression model. This method is a simplified LMIQ method that still employs the basic idea of impersonation. Its performance does not fall much behind that of the full LMIQ method. A comprehensive set of experiments, analyzing the contribution of each type of questionnaire, is detailed in Table~\ref{table:comprehensive_ablation}. The Zero-Shot method that tries to predict the questionnaire's final score directly performs consistently worse.

The classic TF-IDF encoding method obtains the highest MSEs in both tasks. GPT-3.5 Analyze \& Embed and MentaLLama 7B Analyze \& Embed demonstrate mixed performance but consistently surpass the method that directly embeds the interview with Ada-002.

\noindent{\bf Evaluation on the DAIC-WoZ Dataset\quad} 
The WOZ-DAIC dataset \cite{daic} is a predecessor of E-DAIC, which does not include PTSD scores, and in which the interviews contain the questions of a human interviewer. This earlier benchmark provides an opportunity to compare LMIQ with published methods. 

The results of this comparison are presented in Table \ref{table:daic_woz_mae_rmse_comparison} and include both the MAE and RMSE metrics, following the literature. Our method is better than text-based methods on the test set. HCAG~\cite{Niu2021} does not report test set results and is slightly better than us on the development set in one of the two scores. \citet{Dai2021} perform better than us on the development set and worse on the test set, suggesting a possible overfit. \citet{Lu2022} perform worse. We also note that the last two models incorporate audio features, which our approach does not. Finally, we note that \citet{Sadeghi2023}, who also utilized GPT-3.5 Turbo similar to our approach, performed significantly worse on the test set.

\section{Ablating the questionnaires}
LMIQ utilizes questionnaires from four domains: DSM-inspired clinical conditions, Five Factor Model personality traits, therapeutic aspects, and the original questions from the PHQ-8 and PCL-C assessments. By systematically omitting each domain, we analyze their individual and collective contributions to diagnostic accuracy using a RandomForest pipeline.

The results are presented in Table~\ref{table:comprehensive_ablation}. Evidently, the direct questions are the most important for PTSD prediction, where the contribution of other domains is less significant, and they cannot replace the direct questionnaires when these are removed. In the case of Depression, the other three domains (combined) can lead to reasonable results even in the absence of the direct questionnaires and contribute more significantly to the test performance.

\begin{table*}[t]
\centering
\begin{tabular}{@{}ccccccccc@{}}
\toprule
\multicolumn{4}{c}{Questionnaire Domain} &  {\# Features} & \multicolumn{2}{c}{Performance on PHQ}& \multicolumn{2}{c}{Performance on PTSD}\\
\cmidrule{1-4} \cmidrule(lr){6-7} \cmidrule(lr){8-9}
{M. Health} & {Personality} & {Therapuetic} & {Direct} &  & {Dev MSE} & {Test MSE} & {Dev MSE} & {Test MSE}   \\
\midrule
\cmark & \xmark & \xmark & \cmark & 95 & 22.62 & 22.05 & 134.31 & 177.15 \\
\xmark & \xmark & \cmark & \cmark & 40 & 24.60 & 21.00 & 142.25 & 168.62 \\
\xmark & \cmark & \xmark & \cmark & 50 & 26.08 & 20.26 & 143.87 & 166.62 \\
\xmark & \xmark & \xmark & \cmark & 25 & 26.32 & 22.74 & 136.96 & 188.26 \\
\cmark & \cmark & \cmark & \cmark & 135 & 23.87 & 20.42 & 144.17 & 192.93 \\ 
\cmark & \cmark & \xmark & \xmark & 95 & 26.17 & 24.95 & 164.11 & 195.17 \\
\cmark & \xmark & \xmark & \xmark & 70 & 24.88 & 25.44 & 167.94 & 195.81 \\
\cmark & \cmark & \cmark & \xmark & 110 & 25.85 & 23.76 & 173.30 & 230.40 \\
\cmark & \xmark & \cmark & \xmark & 85 & 25.34 & 23.45 & 177.68 & 229.44 \\
\xmark & \cmark & \xmark & \xmark & 25 & 32.13 & 29.47 & 204.92 & 252.67 \\
\xmark & \xmark & \cmark & \xmark & 15 & 32.68 & 26.87 & 238.84 & 255.28 \\
\bottomrule
\end{tabular}
\caption{An ablation study regarding the contribution of the various domains.}
\label{table:comprehensive_ablation}
\end{table*}

\section{Conclusions}
LMIQ employs an LLM as an impersonator to complete both established and preliminary LLM-generated clinical questionnaires based on a subject's unstructured psychological dialogue, feeding these into a classifier for diagnosis. While we do not validate the impersonated responses due to unavailable answers, we demonstrate their effectiveness in predicting PHQ and PCL scores. 
Despite LLMs' limited ability to perform numerical assignments in some contexts, our results affirm the meaningfulness of the assigned values, both directly and through regression trees.

\section{Limitations}

Our method, though generic, has been validated exclusively against the E-DAIC and WOZ-DAIC benchmarks, underscoring a challenge in generalizability. To bridge this gap and encourage clinical adoption, it is essential to amass more data and benchmark our method's performance against the variability observed among different human experts. Such a step is crucial for improving our approach's robustness and applicability in clinical settings, encompassing a broader spectrum of psychological conditions and populations.

The LLM used is a generic one. Experiments conducted with Mixtral 7Bx8 and MentaLLama are limited due to time constraints. It would be interesting to also evaluate Med-PaLM2 and possibly other models. 

An important limitation in our work is the tendency of LLMs to ``hallucinate'' or misinterpret context, a problem exacerbated in the sensitive domain of psychological data where inaccuracies carry significant consequences. Despite this, we aim to identify meaningful connections between unstructured data and symptoms, with the understanding that perfectly mirroring individual experiences is unattainable.

\section{Ethics and Impact Statement}

The use of LMIQ for transforming unstructured text into psychiatric diagnoses presents ethical concerns, notably the possibility of assigning diagnostic labels to individuals without obtaining their explicit consent. Furthermore, the inherent risks associated with misdiagnosis in the field of psychiatry underscore the importance of exercising caution when deploying machine learning tools for these purposes. This caution is necessitated by the recognition of their limited accuracy and the potential for embedded biases, which could lead to harmful outcomes. In alignment with the Health Insurance Portability and Accountability Act (HIPAA) of 1996, our methodology must ensure the de-identification of patient-provider conversations prior to processing with LLMs, safeguarding patient privacy and compliance with legal standards. Therefore, the ethical application of such technologies demands careful consideration of these factors to prevent unintended harm to individuals.

\section{Acknowledgements}
This work was supported by funds from the Sagol Brain Institute, NIMH and the Israel Science Foundation (Grant No. 2923/20) within the Israel Precision Medicine Partnership program. We extend our gratitude for their support.

\bibliography{custom}
\clearpage
\appendix

\section{The full prompts used}
\subsection{System prompt - Question Impersonation}
\label{app:prompt}
This is the prompt used in the main method, LMIQ. 

"""Analyze a therapist-subject conversation transcript and related psychological questionnaire questions. Focus on understanding the subject's mental health by examining their dialogue for both explicit and implicit cues. Pay attention to signs of depression, PTSD, and other conditions, but also note the absence of these symptoms.

Your task is to provide answers to the questionnaire as if you were the subject, based on insights from the conversation. Ensure your responses are balanced, reflecting the subject's mental state as suggested by the transcript. Make informed deductions about the subject's emotional state, stress management, life satisfaction, social interactions, and professional aspirations, providing answers that reflect mental health issues or a neutral state as indicated by the transcript.

Transcript: \{raw\_psyc\_transcript\}\\
Answer the questions with an agreement level ranging from \{agreement\_range\}.\\
Questions: \{questions\}"""

\subsection{System Prompt - Analysis}
\label{app:prompta}
This is the prompt used for the Analyze \& Embed Baseline.

"""Analyze a therapist-subject conversation transcript. Focus on the clear and comprehensible parts of the subject's dialogue, as the therapist's words are omitted and there may be gaps due to recording issues. Your analysis should identify key statements or moments that reveal aspects of the subject's mental health, with particular attention to signs of depression and PTSD. Prioritize brevity and clarity in your analysis. Look for indicators of the subject's emotional state, stress management, life satisfaction, social interactions, and professional aspirations. Assess the intensity of these elements, capturing the range from mild to profound, based on the subject's expressions.

Ensure your analysis remains accurate and relevant to the transcript, avoiding unfounded assumptions or 'hallucinations'. Summarize your findings clearly and directly, using concise sentences that reflect the nuances of the subject's mental state. Disregard unclear or incomplete segments due to recording issues, and avoid structural phrases or introductions to maintain data integrity for subsequent embedding generation. The accuracy, clarity, and nuanced understanding of intensity in your summary are paramount.\\
Conversation transcript: \{raw\_psyc\_transcript\}"""

A sample response using GPT-3.5 is:
\begin{custommdframed}
The subject appears to have a positive outlook on life and their current location. 
They are interested in starting their own business and have a good relationship with their family, who have given them valuable advice. However, there are also signs of stress and irritability, as well as a recent episode of shutting down and cutting off a friend They also mention feeling lazy and less interested in activities they...
Overall, the subject seems to have a mix of positive and negative emotions, and may benefit from further exploration and support for any underlying mental health issues.
\end{custommdframed}

\subsection{Sample output by MentaLLama}
\label{app:promptml}
A sample MentaLLama response is:
\begin{custommdframed}
The subject expresses a positive outlook on life, including enjoyment of reading books, cooking, and spending time with friends and family. However, they also mention experiencing congestion and having a break in their education. They express a desire to open up their own business but do not provide a specific reason. 
The subject mentions receiving advice from their parents and grandparents. 
Overall, the analysis suggests a balance between positive and negative experiences, indicating a potential need for further assessment and support."
\end{custommdframed}

\section{The full list of the questionnaires used.}
\label{app:questions}

For each topic from each domain, we provide the questions that were extracted in collaboration with ChatGPT 4.0.

\subsection{Mental Health Domain}

\begin{itemize}[leftmargin=*,label={}]
    \item \textbf{Agoraphobia}
        \begin{itemize}
            \item Do you fear being in open or enclosed spaces?
            \item Do you avoid public transportation or being in crowds?
            \item Does the thought of leaving your home alone cause you anxiety?
            \item How do these fears limit your daily activities or lifestyle?
            \item Do you require a companion when going out due to these fears?
        \end{itemize}

    \item \textbf{Attention-Deficit/Hyperactivity Disorder}
        \begin{itemize}
            \item Do you find it difficult to organize tasks and activities?
            \item Do you often forget appointments or daily activities?
            \item Do you often make careless mistakes in work or other activities?
            \item Do you get easily sidetracked by extraneous stimuli?
            \item How much do these symptoms impact your performance in work or school?
        \end{itemize}
    \item \textbf{Body Image}
        \begin{itemize}
            \item How would you describe your level of satisfaction with your appearance?
            \item Do you often compare your body to others?
            \item How much does your perception of your body affect your daily life?
            \item Do you have any concerns about your eating habits or weight management?
            \item How confident do you feel in your abilities and decisions?
        \end{itemize}
    \item \textbf{Borderline Personality Disorder}
        \begin{itemize}
            \item Do you experience intense and unstable relationships with others?
            \item Do you often feel empty or bored?
            \item Do you experience mood swings that can last for a few hours to a few days?
            \item Do you have a fear of abandonment, either real or imagined?
            \item Do you engage in impulsive behaviors, like substance abuse or reckless driving?
        \end{itemize}
    \item \textbf{Delusions}
        \begin{itemize}[font=\bfseries] 
            \item Do you feel disconnected from reality at times?
            \item Do you have strong beliefs that others find unusual or unrealistic?
            \item Do you feel controlled or influenced by external forces or beings?
            \item Have you experienced changes in your perception or senses that others do not?
            \item Do these experiences cause you distress or impair your functioning?
        \end{itemize}
    \item \textbf{Generalized Anxiety Disorder}
        \begin{itemize}
            \item Do you find it hard to control your worrying?
            \item Does your anxiety interfere with your work, school, or family responsibilities?
            \item Do you experience physical symptoms of anxiety, like muscle tension or restlessness?
            \item Do you often feel irritable or on edge?
            \item Do you have trouble sleeping due to worry?
        \end{itemize}

    \item \textbf{Hypomania/Mania}
        \begin{itemize}
            \item Have you found yourself more talkative or speaking faster than usual?
            \item Do you often feel overly confident in your abilities or ideas?
            \item Have you engaged in risky behaviors, like excessive spending or reckless driving?
            \item Do you find your thoughts racing or jumping from topic to topic?
            \item Have others noticed a significant change in your mood or behavior?
        \end{itemize}
    \item \textbf{Major Depressive Disorder}
        \begin{itemize}
            \item Do you often feel hopeless or helpless?
            \item Have you noticed a change in your appetite or weight without trying to lose or gain weight?
            \item Do you struggle to concentrate on tasks or make decisions?
            \item Do you often feel worthless or excessively guilty about things?
            \item Have your sleep patterns changed, such as sleeping too much or too little?
        \end{itemize}
    \item \textbf{Obsessive-Compulsive Disorder}
        \begin{itemize}
            \item Do you check things repeatedly or have rituals that you feel compelled to perform?
            \item Do your thoughts or rituals cause you distress or interfere with your daily life?
            \item Do you spend more than an hour a day on these thoughts or rituals?
            \item Do you avoid certain situations or activities because of your fears or compulsions?
            \item Do you need to have things arranged in a specific order or manner?
        \end{itemize}
        \item \textbf{Panic Attacks}
        \begin{itemize}
            \item During panic attacks, do you feel like you're losing control or going crazy?
            \item Do you fear these attacks to the point of altering your daily routines?
            \item Have you visited the emergency room or sought medical help for these symptoms?
            \item Do you avoid places or situations for fear of triggering an attack?
            \item How do these attacks impact your daily life?
        \end{itemize}

    \item \textbf{Persistent Depressive Disorder}
        \begin{itemize}
            \item Have you experienced low mood more days than not for at least two years?
            \item Do you feel like you've been in a mild but constant state of depression?
            \item Do you find little pleasure in activities you once enjoyed?
            \item Do you struggle with feelings of inadequacy or low self-esteem?
            \item Have you experienced changes in your appetite or sleep patterns?
        \end{itemize}

    \item \textbf{Post-Traumatic Stress Disorder}
        \begin{itemize}
            \item Do you experience heightened vigilance or jumpiness?
            \item Are you engaging in self-destructive or risky behavior since the event?
            \item Do you feel numb or detached from people, activities, or surroundings?
            \item Do you find yourself being easily angered or having aggressive outbursts?
            \item Have you noticed any changes in your beliefs or feelings about yourself and others?
        \end{itemize}

    \item \textbf{Social Phobia}
        \begin{itemize}
            \item Do you fear being criticized or embarrassed in social situations?
            \item Does speaking to unfamiliar people cause you significant anxiety?
            \item Do you avoid social situations due to fear of being judged?
            \item Do physical symptoms like sweating or trembling accompany your fear in social settings?
            \item How does this fear impact your personal or professional life?
        \end{itemize}

    \item \textbf{Substance Abuse}
        \begin{itemize}
            \item Do you use any substances like drugs or alcohol regularly?
            \item How often do you find yourself using these substances?
            \item Have you noticed any negative impacts on your health, work, or relationships due to substance use?
            \item Do you feel a strong desire or compulsion to use these substances?
            \item Have you tried to cut down or stop using these substances in the past?
        \end{itemize}

\end{itemize}
\subsection{Personality}
\begin{itemize}[leftmargin=*,label={}]
    \item \textbf{Agreeableness}
        \begin{itemize}
            \item Do you often find yourself making compromises to maintain harmony in your relationships?
            \item Would you describe yourself as someone who is generally trusting of others?
            \item How often do you get into arguments with people?
            \item Do you tend to empathize easily with others?
        \end{itemize}
    \item \textbf{Conscientiousness}
        \begin{itemize}
            \item How often do you set and achieve long-term goals?
            \item Do you prefer having a set schedule or being spontaneous?
            \item How would you rate your ability to resist temptations or distractions?
            \item Do you take pride in the accuracy and detail of your work?
            \item How do you handle important deadlines?
        \end{itemize}
    \item \textbf{Extraversion}
        \begin{itemize}
            \item Do you enjoy being the center of attention in social gatherings?
            \item How often do you initiate conversations with people you don’t know?
            \item Do you prefer group activities or solitary activities?
            \item Do you feel energized when interacting with a large group of people?
            \item How would you describe your level of assertiveness in social situations?
        \end{itemize}
    \item \textbf{Neuroticism}
        \begin{itemize}
            \item Do you often feel anxious or worried about various aspects of your life?
            \item How do you react to stressful situations?
            \item Do you frequently feel mood swings or emotional instability?
            \item Do you often have trouble sleeping due to worrying?
            \item How often do you experience feelings of sadness or depression?
        \end{itemize}
    \item \textbf{Openness}
        \begin{itemize}
            \item Do you enjoy trying new activities and visiting new places?
            \item How often do you engage in creative activities like writing, painting, or playing music?
            \item Do you enjoy discussing abstract concepts and ideas?
            \item How do you feel about change and variety in your life?
            \item Would you say you are open to new and diverse perspectives or opinions?
        \end{itemize}
\end{itemize}

\subsection{Therapeutic}
\begin{itemize}[leftmargin=*,label={}]
    \item \textbf{Family History}
        \begin{itemize}
            \item Do you feel you have a strong and positive relationship with your family members?
            \item Do you manage conflicts or disagreements within your family effectively?
            \item Is there a history of mental health issues or substance use in your family?
            \item Do you believe your family background has significantly influenced your current life choices and behaviors?
            \item Do you feel supported and understood by your family?
        \end{itemize}
    \item \textbf{Trauma History}
        \begin{itemize}
            \item Have you ever experienced a traumatic event such as physical, emotional, or sexual abuse?
            \item How do you feel this event has affected your life?
            \item Do you often think about or have flashbacks to this traumatic event?
            \item How do you typically cope with reminders of the trauma?
            \item Have you sought any professional help to deal with the aftermath of this traumatic experience?
        \end{itemize}
    \item \textbf{Resilience}
        \begin{itemize}
            \item How quickly do you recover from setbacks or disappointments?
            \item Do you often find positive aspects in negative situations?
            \item How do you usually cope with stress and pressure?
            \item Do you feel confident in your ability to handle new challenges?
            \item How often do you bounce back from hardships stronger than before?
        \end{itemize}
\end{itemize}
\subsection{Direct}
\begin{itemize}[leftmargin=*,label={}]
    \item \textbf{PHQ-8}
        \begin{itemize}
            \item How much are you experiencing little interest or pleasure in doing things?
            \item How likely are you to volunteer your time to help others?
            \item How much are you feeling down, depressed, or hopeless?
            \item How much are you having trouble with falling or staying asleep, or sleeping too much?
            \item How much are you feeling tired or having little energy?
            \item How much are you experiencing poor appetite or overeating?
            \item How much are you feeling bad about yourself, or that you are a failure or have let yourself or your family down?
            \item How much are you having trouble concentrating on things, like reading the newspaper or watching television?
            \item How much are you moving or speaking so slowly that other people might have noticed, or the opposite – being so fidgety or restless that you've been moving around a lot more than usual?
        \end{itemize}
    \item \textbf{PCL-C}
        \begin{itemize}
            \item How much are you re-experiencing disturbing memories, thoughts, or images of a stressful experience from the past?
            \item How much are you experiencing repeated, disturbing dreams of a stressful experience from the past?
            \item How much are you suddenly acting or feeling as if a stressful experience were happening again (as if you were reliving it)?
            \item How much are you feeling upset when something reminded you of a stressful experience from the past?
            \item How much are you having physical reactions (e.g., heart pounding, trouble breathing, sweating) when something reminded you of a stressful experience?
            \item How much are you avoiding thinking about or talking about a stressful experience from the past or avoiding having feelings related to it?
            \item How much are you avoiding activities or situations because they remind you of a stressful experience?
            \item How much are you having trouble remembering important parts of a stressful experience?
            \item How much are you losing interest in activities that you used to enjoy?
            \item How much are you feeling detached or estranged from others?
            \item How much are you feeling emotionally numb or being unable to have loving feelings for those close to you?
            \item How much are you feeling as if your future will somehow be cut short?
            \item How much are you having trouble falling or staying asleep?
            \item How much are you feeling irritable or having angry outbursts?
            \item How much are you having difficulty concentrating?
            \item How much are you being 'super alert' or watchful or on guard?
            \item How much are you feeling jumpy or easily startled?
        \end{itemize}
\end{itemize}

\clearpage

\section{Interpretability Analysis of LMIQ}

A useful property of the Random Forest algorithm is the ability to identify the most important features. Applying this feature importance analysis to the models used to predict PHQ Scores and PCL Severity is provided in Tables ~\ref{table:phqImportance} and \ref{table:pclImportance}, respectively, revealing the questions that the LMIQ models predominantly rely on for predictions. 


The results 
show a strong alignment between questions, the associated symptoms,  and their respective domains (depression or PTSD), highlighting the model's promise. The inclusion of less obvious questions among those deemed highly relevant, such as "Do you often find positive aspects in negative situations?" from the PHQ score influences, and "Do you experience mood swings that can last for a few hours to a few days?" from the PCL severity influences, underscores an opportunity to further explore the model's internal reasoning and its capacity to link everyday language with a wide range of symptoms and behaviors. 

\begin{table}
\centering
\begin{tabular}{ll}
\toprule
Feature & Relative Importance \\
\midrule
Do you often have trouble sleeping due to worrying & 0.18 \\
Have your sleep patterns changed, such as sleeping too much or too little? & 0.16 \\
Do these experiences cause you distress or impair your functioning? & 0.03 \\
Do you often find positive aspects in negative situations? & 0.03 \\
How often do you experience feelings of sadness or depression? & 0.02 \\
\bottomrule
\end{tabular}
\caption{Top 5 Questions Influencing the PHQ Score}
\label{table:phqImportance}
\end{table}
\begin{table}
\begin{tabular}{ll}
\toprule
Feature & Relative Importance \\
\midrule
During panic attacks, do you feel like you're losing control or going crazy? & 0.09 \\
How do these attacks impact your daily life? & 0.08 \\
How do these fears limit your daily activities or lifestyle? & 0.06 \\
Do your thoughts or rituals cause you distress or interfere with your daily life? & 0.04 \\
Do you experience mood swings that can last for a few hours to a few days? & 0.03 \\
\bottomrule
\end{tabular}
\centering
\caption{Top 5 Questions Influencing the PCL Severity}
\label{table:pclImportance}
\end{table}

\label{spp:questions}
\end{document}